\documentclass[11pt,a4paper]{article}
\usepackage[hyperref]{acl2018}
\usepackage{times}
\usepackage{latexsym}

 \usepackage{float}
\usepackage{url}
\usepackage{array, caption, tabularx, makecell, booktabs}%
\usepackage{multirow}
\usepackage{color}
\usepackage{colortbl}
\usepackage[colorlinks]{}
\usepackage{subcaption}
\usepackage{graphicx}
\usepackage{amsmath}
\usepackage{breqn}
\usepackage [autostyle, english = american]{csquotes}
\MakeOuterQuote{"}

\aclfinalcopy 

\title{Can Neural Generators for Dialogue \\ Learn Sentence Planning and Discourse Structuring?}

\author{Lena Reed, Shereen Oraby and Marilyn Walker\\
Natural Language and Dialogue Systems Lab, University of California, Santa Cruz\\
  {\tt \{lireed,soraby,mawalker\}@ucsc.edu} \\}

\date{}

\begin{document}
\maketitle
\begin{abstract}
Responses in task-oriented dialogue systems often realize multiple
propositions whose ultimate form depends on the use of sentence
planning and discourse structuring operations. For example a
recommendation may consist of an explicitly evaluative utterance
e.g. {\it Chanpen Thai is the best option}, along with content related
by the justification discourse relation, e.g. {\it It has great food
  and service}, that combines multiple propositions into a
single phrase.  While neural generation methods integrate
sentence planning and surface realization in one end-to-end learning
framework, previous work has not shown that neural
generators can: (1) perform common sentence planning and discourse
structuring operations;
(2) make decisions as to whether to
realize content in a single sentence or over multiple sentences;
(3) generalize sentence planning and discourse relation operations 
beyond  what was seen in training.
We systematically create large training corpora that exhibit
particular sentence planning operations and then test neural models to
see what they learn.  We compare models without explicit latent
variables for sentence planning with ones that provide explicit
supervision during training. We show that only the models with
additional supervision can reproduce sentence planning and discourse
operations and generalize to situations unseen in
training.
\end{abstract}
\vspace{-.2in}
\section{Introduction}
\label{intro-sec}

Neural natural language generation ({\sc nnlg}) promises to simplify
the process of producing high quality responses for conversational
agents by relying on the neural architecture to automatically learn
how to map an input meaning representation (MR) from the dialogue manager
to an output utterance \cite{gavsic2017spoken,sutskever2014sequence}.
For example, Table~\ref{table:scope-examples} shows sample 
training data for an {\sc nnlg} with a 
MR for a restaurant named {\sc zizzi}, along with
three reference realizations, that should allow the {\sc nnlg} to learn
to realize the MR as either 1, 3, or 5 sentences.
\begin{table}[h!t]
\begin{small}
\begin{tabular}
{@{} p{0.01in}|p{0.4in}|p{2.25in}@{}}
\toprule
{\bf \#} & {\bf Type} & {\bf  Example} \\ \hline                 
\rowcolor[gray]{0.9}\multicolumn{3}{l}{{\sc priceRange[moderate], area[riverside],}} \\
\rowcolor[gray]{0.9}\multicolumn{3}{l}{{\sc name[Zizzi], food[English], eatType[pub] }}  \\
\rowcolor[gray]{0.9}\multicolumn{3}{l}{ {\sc near[Avalon], familyFriendly[no] }}   \\       \hline                  
1 & 1 Sent &   Zizzi is moderately priced in riverside, also it isn't family friendly, also it's a pub, and it is an English place near Avalon. \\ \hline   
2 & 3 Sents & Moderately priced Zizzi isn't kid friendly, it's in riverside and it is near Avalon. It is a pub. It is an English place. \\ \hline   
3 & 5 Sents &  Zizzi is moderately priced near Avalon. It is a pub. It's in riverside. It isn't family friendly. It is an English place. \\ \hline   
\end{tabular}
\end{small}
\vspace{-.1in}
\centering \caption{\label{table:scope-examples} {Sentence Scoping: a sentence planning operation that decides what content to place in each sentence of an utterance.}} 
\end{table}

\begin{figure}[!ht]
\centering
\includegraphics[width=2.9in]{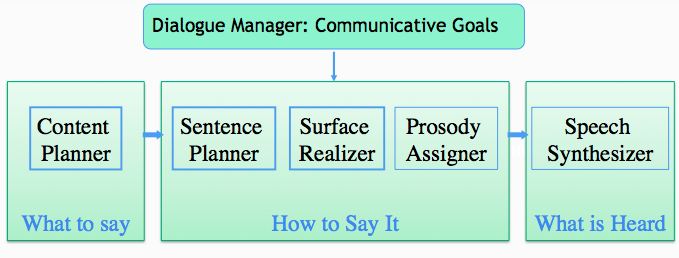}
\vspace{-.2in}
\caption{\label{snlg-arch} Statistical NLG Dialogue Architecture}
\end{figure}

In contrast, earlier models of statistical natural language generation
({\sc snlg}) for dialogue were based around the NLG
architecture in Figure~\ref{snlg-arch} \cite{RRW01,Stent02,stentmolina09}. 

Here the dialogue manager sends one or more dialogue acts and their
arguments to the NLG engine, which then makes decisions how to render
the utterance using separate modules for content planning and
structuring, sentence planning and surface realization
\cite{reiter00book}.  The sentence planner's job includes:

\begin{itemize}  \setlength\itemsep{-0.2em}
\item {\bf Sentence Scoping}: deciding how to allocate the content to be expressed across different sentences;
\item {\bf Aggregation}: implementing strategies for
removing redundancy and constructing compact sentences;
\item {\bf Discourse Structuring}: deciding how to express discourse relations that hold between
content items, such as causality, contrast, or justification. 
\end{itemize}

Sentence scoping (Table~\ref{table:scope-examples})
affects the complexity of
the sentences that compose an utterance, allowing the NLG to produce
simpler sentences when desired that might be easier for particular
users to understand.  Aggregation reduces redundancy, composing
multiple content items into single sentences. 
Table~\ref{table:agg-examples} shows common aggregation
operations \cite{cahilletalacl01,shaw98}.  Discourse structuring is
often critical in persuasive settings 
\cite{ScottSouza90,MooreParis93}, in order to express
discourse relations that hold between content items.
Table
~\ref{table:discourse-examples} shows how {\sc recommend}
dialogue acts can be included in the MR, and how content can be
related with {\sc justify} and {\sc contrast} discourse
relations \cite{Stentetal02}.

Recent work in {\sc nnlg} explicitly claims that training models
end-to-end allows them to do {\bf both} sentence planning and surface
realization without the need for intermediate 
representations
\cite{DusekJ16,Lampouras2016,Mei2015,Wenetal15,Nayaketal17}.  To date,
however, no-one has actually shown that an {\sc nnlg} can faithfully
produce outputs that exhibit the sentence planning and discourse
operations in Tables~\ref{table:scope-examples},
~\ref{table:agg-examples} and ~\ref{table:discourse-examples}.
Instead, {\sc nnlg} evaluations  focus 
on measuring the semantic correctness of the outputs and their
fluency \cite{NovikovaetalEVAL17,Nayaketal17}.

\begin{table}[t!]
\vspace{-.1in}
\begin{small}
\begin{tabular}
{@{} p{0.01in}|p{0.55in}|p{2.1in}@{}}
\toprule
{\bf \#} & {\bf Type} & {\bf  Example} \\ \hline                 
\rowcolor[gray]{0.9}\multicolumn{3}{l}{ {\sc name[The Mill], eatType[coffee shop], }} \\
 \rowcolor[gray]{0.9}\multicolumn{3}{l}{ {\sc food[Italian], priceRange[low],}}  \\      
\rowcolor[gray]{0.9}\multicolumn{3}{l}{ {\sc  customerRating[high], near[The Sorrento]}}  \\       \hline   
4 & With, Also &  The Mill is a coffee shop with a high rating with a low cost, also The Mill is an Italian place near The Sorrento. \\ \hline         
5 & With, And &  The Mill is a coffee shop with a high rating with a high cost and it is an Italian restaurant near The Sorrento. \\ \hline   
6 & Distributive &  The Mill is a coffee shop with a {\bf high rating and cost}, also it is an Italian restaurant near The Sorrento. \\ \hline   
\end{tabular}           
\end{small}
\vspace{-.1in}
\centering \caption{\label{table:agg-examples} {Aggregation Operation Examples}} 
\end{table}

\begin{table}[t!]
\begin{small}
\begin{tabular}
{@{} p{0.01in}|p{0.81in}|p{1.8in}@{}}
\toprule
{\bf \#} & {\bf Discourse Rel'n} & {\bf  Example} \\ \hline                 
\rowcolor[gray]{0.9} \multicolumn{3}{l}{ {\sc name[Babbo], recommend[yes], }} \\
\rowcolor[gray]{0.9}\multicolumn{3}{l}{ {\sc
food[Italian], price[cheap],}}  \\       
\rowcolor[gray]{0.9}\multicolumn{3}{l}{ {\sc  qual[excellent], near[The Sorrento], }} \\
\rowcolor[gray]{0.9}\multicolumn{3}{l}{ {\sc location[West Village], service[poor]}} \\ \hline   
7 & {\sc justify ([recommend] [food, price, qual])}  & I would suggest Babbo {\bf because it serves 
Italian food with excellent quality and it is inexpensive}. 
The service is poor and it is near the Sorrento in the West Village. \\ \hline  
8 & {\sc contrast [price, service]}   & I would suggest Babbo because it serves 
Italian food with excellent quality and {\bf it is inexpensive. However the service is poor}. It is near the Sorrento in the West Village. \\ \hline  
\end{tabular}           
\end{small}
\vspace{-.1in}
\centering \caption{\label{table:discourse-examples} {Justify \& Contrast Discourse Relations}} 
\end{table}

Here, we systematically perform a set of
controlled experiments to test whether an {\sc nnlg} can learn to do
sentence planning operations.
Section~\ref{models-sec} describes our experimental setup and the {\sc
  nnlg} architecture that allows us, during training, to vary the
amount of supervision provided as to which sentence planning
operations appear in the outputs. To ensure that the training data
contains enough examples of particular phenomena, we experiment with
supplementing crowdsourced data with automatically generated
stylistically-varied data from {\sc personage}
\cite{MairesseWalker11}.  To achieve sufficient control for some
experiments, we exclusively use Personage training data where we can
specify exactly which sentence planning operations will be used and in
what frequency. It is not possible to do this with crowdsourced
data. While our expectation was that an {\sc nnlg} can reproduce any
sentence planning operation that appears frequently enough in the
training data, the results in Sections~\ref{scope-results-sec},
~\ref{distrib-results-sec} and ~\ref{contrast-results-sec} show that
explicit supervision improves the {\bf semantic accuracy} of the {\sc
  nnlg}, provides the capability to {\bf control} variation in the
output, and enables {\bf generalizing} to unseen value combinations.

\section{Model Architecture and Experimental Overview}
\label{models-sec}

Our experiments focus on sentence planning operations for:  (1)
  sentence scoping, as in Table~\ref{table:scope-examples},
where we experiment with controlling the number of sentences in the
generated output;  (2) distributive aggregation, as in Example 6
in Table~\ref{table:agg-examples}, an aggregation operation that
can compactly express a description when two attributes share the same
value; 
and (3) discourse contrast, as in Example 8 in
Table~\ref{table:discourse-examples}.

Distributive aggregation requires learning a proxy for the semantic
property of {\bf equality} along with the standard mathematical {\bf
  distributive} operation, while discourse contrast requires learning
a proxy for  semantic comparison, i.e. that some attribute values are
evaluated as positive ({\it inexpensive}) while others are evaluated
negatively ({\it poor service}), and that a successful contrast can
only be produced when two attributes are on opposite poles (in either order), as defined
in Figure~\ref{sem-learning-fig}.\footnote{We also note that the evaluation of an attribute may come from the attribute itself, e.g. "kid friendly", or from its adjective, e.g. "excellent service".}

\begin{figure}[h]
\begin{small}
\begin{center}
\begin{tabular}[h]
{@{} |p{.1in} p{2.0in}|@{}}
\hline
\rowcolor[gray]{0.9} \multicolumn{2}{|l|} {{\sc Distributive Aggregation}} \\
\multicolumn{2}{|l|} 
{{if $ATTR_1$ := $ADJ_i$}} \\ 
 & {and} $ATTR_2$ := $ADJ_j$ \\ 
 & {and} $ADJ_i$ = $ADJ_j$ \\ 
\multicolumn{2}{|l|} {then DISTRIB($ATTR_1,ATTR_2$)} \\
& \\
\rowcolor[gray]{0.9} \multicolumn{2}{|l|} {{\sc Discourse Contrast}} \\
\multicolumn{2}{|l|} 
{{if EVAL($ADJ_i(ATTR_1)$) = POS}} \\ 
 & {and} EVAL($ADJ_j(ATTR_2)$) = NEG \\ 
\multicolumn{2}{|l|} {then CONTRAST($ATTR_1,ATTR_2$)} \\
\hline
\end{tabular}
\end{center}
\end{small}
\caption{Semantic operations underlying distributive aggregation and 
contrast \label{sem-learning-fig}}
\end{figure}

\begin{figure}[t!h]
\centering
\includegraphics[width=\linewidth, keepaspectratio]{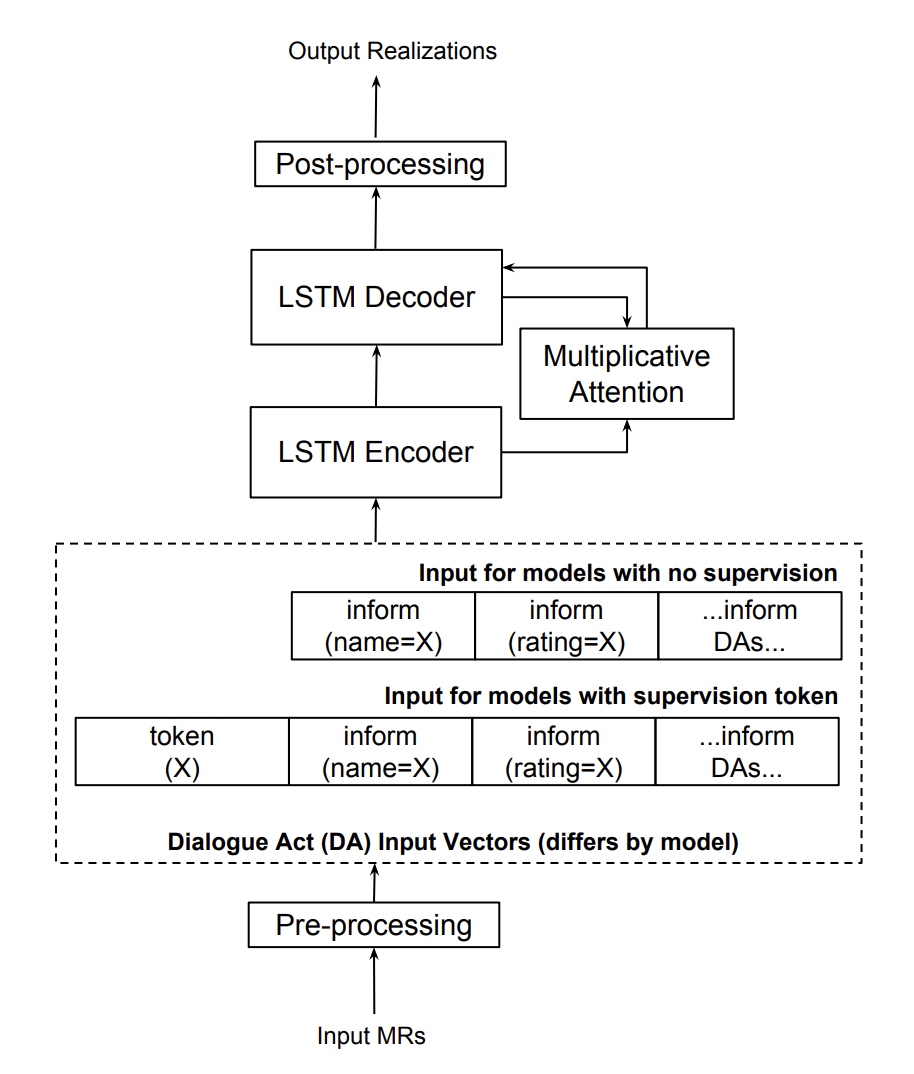}
\vspace{-.25in}
\caption{\label{fig:model1}Neural Network Model Architecture, illustrating both the {\sc no supervision} baseline and models that add the {\sc token} supervision}
\end{figure}

Our goal is to test how well {\sc nnlg} models can produce
realizations of these sentence planning operations with
varying levels of supervision, while simultaneously achieving
high semantic fidelity.  Figure \ref{fig:model1} shows the
general architecture, implemented in Tensorflow, based on TGen, an
open-source sequence-to-sequence (seq2seq) neural generation framework
\cite{Abadi16,DusekJ16a}.\footnote{\url{https://github.com/UFAL-DSG/tgen}}
The model uses seq2seq generation with attention
\cite{bahdanau2014neural, sutskever2014sequence} with a
sequence of LSTMs \cite{hochreiter1997long} for encoding and decoding,
along with beam-search and an n-best reranker.

The input to the sequence to sequence model is a sequence of tokens
$x_{t}, t \in \{0, \ldots, n\}$ that represent the dialogue act and
associated arguments. Each $x_{i}$ is associated with an embedding
vector $w_{i}$ of some fixed length.  Thus for each MR, TGen takes as
input the dialogue acts representing system actions ({\it recommend}
and {\it inform} acts) and the attributes and their values (for
example, an attribute might be {\it price range}, and its value might
be {\it moderate}), as shown in Table \ref{table:scope-examples}.  The
MRs (and resultant embeddings) are sorted internally by dialogue act
tag and attribute name. For every MR in training, we have a matching
reference text, which we delexicalize in pre-processing, then
re-lexicalize in the generated outputs.  The encoder reads all the
input vectors and encodes the sequence into a vector
$h_n$. At each time step $t$, it computes the hidden
layer $h_t$ from the input $w_t$ and hidden vector at the previous
time step $h_{t-1}$, following:

	\[h_t = (W_1 . x_t + W_2 . h_{t-1}) + b\]

All experiments use a standard LSTM decoder.

We test three different
dialogue act and input vector representations, based on the level of
supervision, as shown by the two input vectors in
Figure~\ref{fig:model1}: {\bf (1) models with no supervision}, where
the input vector simply consists of a set of {\it inform} or {\it
  recommend} tokens each specifying an attribute and value pair, and
{\bf (2) models with a supervision token}, where the input vector is
supplemented with a new token (either {\it period} or {\it distribute}
or {\it contrast}), to represent a latent variable to guide the {\sc
  nnlg} to produce the correct type of sentence planning operation;
{\bf (3) models with semantic supervision}, tested only on distributive
aggregation, where the input vector is supplemented with specific
instructions of which attribute value to distribute over, e.g. {\it
  low}, {\it average} or {\it high}, in the {\sc
  distribute} token.  We describe the specific model variations for
each experiment below.

\noindent{\bf Data Sets.} One challenge is that {\sc nnlg} models are
highly sensitive to the distribution of phenomena in training data,
and our previous work has shown that the outputs of {\sc nnlg} models
exhibit less stylistic variation than their training data
\cite{Orabyetal18}.  Moreover, even large corpora, such as the 50K E2E
Generation Challenge corpus, may not contain particular stylistic
variations. For example, out of 50K crowdsourced examples in the E2E
corpus, there are 1,956 examples of contrast with the operator
"but". There is only 1 instance of distributive aggregation because
attribute values are rarely lexicalized identically in E2E.  To ensure
that the training data contains enough examples of particular
phenomena, our experiments combine crowdsourced E2E
data\footnote{{\url{http://www.macs.hw.ac.uk/InteractionLab/E2E/}}}
with automatically generated data from {\sc personage}
\cite{MairesseWalker11}.\footnote{Source code for {\sc personage} was provided by Fran\c{c}ois
  Mairesse.}  This allows us to systematically create training data
that exhibits particular sentence planning operations, or combinations
of them. The E2E dataset consists of pairs of reference utterances and
their meaning representations (MRs), where each utterance contains up
to 8 unique attributes, and each MR has multiple references.  We
populate {\sc personage} with the syntax/meaning mappings that it
needs to produce output for the E2E meaning representations, and then
automatically produce a very large (204,955 utterance/MR pairs)
systematically varied sentence planning corpus.\footnote{We make
available the sentence
planning for NLG corpus at: nlds.soe.ucsc.edu/sentence-planning-NLG.}

{\noindent{\bf Evaluation metrics}. It is well known that evaluation
  metrics used for translation such as BLEU are not well suited to
  evaluating generation outputs
  \cite{BelzReiter06,Liuetal16b,NovikovaetalEVAL17}: they penalize
  stylistic variation, and don't account for the fact that different
  dialogue responses can be equally good, and can vary due to
  contextual factors \cite{jordan-cogsci2k,Krahmer2002}.  We also note
  that previous work on sentence planning has always assumed that
  sentence planning operations improve the quality of the output
  \cite{BarzilayLapata06,shaw98}, while our primary focus here is to
  determine whether an {\sc nnlg} can be trained to perform such
  operations while maintaining semantic fidelity.  Moreover, due to
  the large size of our controlled training sets, we observe few
  problems with output quality and fluency.

Thus we leave an evaluation of fluency and naturalness to future work,
and focus here on evaluating the multiple targets of semantic accuracy
and sentence planning accuracy. Because the MR is clearly defined, we
define scripts (information extraction patterns) to measure the
occurrence of the MR attributes and their values in the outputs. We
then compute Slot Error Rate (SER) using a variant of word error rate:
\begin{center}
${\displaystyle {\mathit {SER}}={\frac {S+D+I+H}{N}}}$
\end{center}
where $S$ is the number of substitutions, $D$ is the number of deletions,
$I$ is the number of insertions, $H$ is the number of hallucinations and $N$ is the number of 
slots in the input MR. 

We also define scripts for evaluating the accuracy of the sentence
planner's operations. We check whether: (1) the output has the right
number of sentences; (2) attributes with equal values are realized
using distributive aggregation, and (3) discourse contrast is used
when semantically appropriate.  Descriptions of each experiment
and the results are in Section~\ref{scope-results-sec},
Section~\ref{distrib-results-sec}, and
Section~\ref{contrast-results-sec}.

\section{Sentence Scoping Experiment}
\label{scope-results-sec}

To test whether it is possible to control basic sentence scoping with
an {\sc nnlg}, we experiment first with controlling the number of
sentences in the generated output, as measured using the period
operator. See Table \ref{table:scope-examples}.  We experiment with
two different models:
\begin{itemize}  \setlength\itemsep{-0.2em}
\item {\noindent \bf No Supervision:} no additional information in the MR (only attributes and their values)
\item {\noindent \bf Period Count Supervision:} has an additional supervision token, {\sc period}, specifying the number of periods (i.e. the number of sentences) to be used in the output realization.
\end{itemize}

For sentence scoping, we construct a training set of 64,442 output/MR
pairs and a test set of 398 output/MR pairs where the reference
utterances for the outputs are generated from {\sc
  personage}. Table~\ref{period-mr} shows the number of training
instances for each MR size for each period count.  The right frontier
of the table shows that there are low frequencies of training
instances where each proposition in the MR is realized in its own
sentence (Period = Number of MR attrs -1). The lower left hand side of
the table shows that as the MRs get longer, there are lower
frequencies of utterances with Period=1.

 \begin{table}[H]
 \centering
  \begin{small}
 \begin{tabular}{|c|l|c|c|c|c|c|c|c|}
 \hline
 & & \multicolumn{7}{c|}{\bf Number of Periods}\\
 \hline
 & \multicolumn{1}{c|}{} & \multicolumn{1}{c|}{\bf 1} & \multicolumn{1}{c|}{\bf 2} & \multicolumn{1}{c|}{\bf 3} & \multicolumn{1}{c|}{\bf 4} & \multicolumn{1}{c|}{\bf 5} & \multicolumn{1}{c|}{\bf 6} & \multicolumn{1}{c|}{\bf 7}\\
 \hline
 \parbox[t]{2mm}{\multirow{6}{*}{\rotatebox[origin=c]{90}{\bf Attributes}}} & \bf 3 &3745&167&0&0&0&0&0\\
 & \bf 4 &5231&8355&333&0&0&0&0\\
 & \bf 5 &2948&9510&7367&225&0&0&0\\
 & \bf 6 &821&5002&7591&3448&102&0&0\\
 & \bf 7 &150&1207&2983&2764&910&15&0\\
 & \bf 8 &11&115&396&575&388&82&1\\
 \hline
 \end{tabular}
 \caption{\label{period-mr} Distribution of Training Data}
  \end{small}
 \end{table}


We start with the default TGen parameters and monitor the losses on
Tensorboard on a subset of 3,000 validation instances from the 64,000
training set. The best settings use a batch size of 20, with a minimum
of 5 epochs and a maximum of 20 (with early-stopping based on
validation loss). We generate outputs on the test set of 398 MRs.

\noindent{\bf Sentence Scoping Results.}  Table
\ref{table:period-results} shows the accuracy of both models in terms
of the counts of the output utterances that realize the MR attributes
in the specified number of sentences. In the case of {\sc
NoSup}, we compare the number of sentences in the generated
output to those in the corresponding test reference, and for {\sc
PeriodCount}, we compare the number of sentences in the
generated output to the number of sentences we {\it explicitly} encode
in the MR. The table shows that the {\sc NoSup} setting fails to
output the correct number of sentences in most cases (only a 22\%
accuracy), but the {\sc PeriodCount} setting makes only 2 mistakes
(almost perfect accuracy), demonstrating almost perfect control of the
number of output sentences with the single-token supervision. We also
show correlation levels with the gold-standard references (all
correlations significant at $p \le 0.01$).

\begin{table}[!htb]
\begin{small}
\begin{tabular}
{@{} p{0.85in}|p{0.47in}|p{0.53in}|p{0.63in}@{}}
\hline
\bf  Model & \bf Slot  & \bf Period & \bf Period  \\ 
           & \bf Error & \bf Accuracy & \bf Correlation \\ \midrule
{\sc NoSup }       &  .06    & 0.216                   &  0.455 \\ \hline
{\sc Period Count  } & .03  & 0.995                   & 0.998   \\ \bottomrule
\end{tabular}
\end{small}
\centering \caption{\label{table:period-results} Sentence Scoping Results}
\end{table}

\noindent{\bf Generalization Test.} We carry out an additional
experiment to test generalization of the {\sc PeriodCount} model, where
we randomly select a set of 31 MRs from the test set, then create a set
instance for each possible {\sc period} count value, from 1 to the
N-1, where N is the number of attributes in that MR (i.e. {\sc
  period=1} means all attributes are realized in the same sentence,
and {\sc period=N-1} means that each attribute is realized in its own
sentence, except for the restaurant name which is never realized in
its own sentence). This yields 196 MR and reference pairs.

This experiment results in an 84\% accuracy (with correlation of 0.802
with the test refs, $p \le 0.01$). When analyzing the mistakes, we
observe that all of the scoping mistakes the model makes (31 in total)
are the case of {\sc period=N-1}.  These cases correspond to the right
frontier of Table~\ref{period-mr} where there were fewer
training instances. Thus while the period supervision improves the
model, it still fails on cases where there were few instances in
training.

\noindent{\bf Complexity Experiment.} We performed an additional
sentence scoping experiment where we specified a target sentence
complexity instead of a target number of sentences, since this may
more intuitively correspond to a notion of reading level or sentence
complexity, where the assumption is that longer sentences are more
complex \cite{Howcroftetal17,Siddharthanetal04}.  We used the same
training and test data, but labeled each reference
as either high, medium or low complexity. The number of attributes in
the MR does not include the name attribute, since that is the subject
of the review. A reference was labeled high when there are $>2$
attributes per sentence, medium when the number of attributes per
sentence is $>1.5$ and $\le 2$ and low when there are $\le 1.5$
attributes per sentence. 

This experiment results in 89\% accuracy. Most of the errors occur
when the labeled complexity was medium. This is most likely because
there is often only one sentence difference between the two complexity
labels.  This indicates that sentence scoping can be used to create
references with either exactly the number of sentences requested or
categories of sentence complexity.

\section{Distributive Aggregation Experiment}
\label{distrib-results-sec}

\begin{table}[htb!]
\centering 
\begin{footnotesize}
\begin{tabular}
{@{} p{1.2in}|p{1.65in} @{}}
\hline
{\bf Operation} & {\bf Example} \\ \hline
{\sc Period} &  {\it X serves Y. It is in  Z.}  \\
{\sc "With" cue} &  {\it X is in Y, with Z.}  \\
{\sc Conjunction} & {\it X is Y and it is Z. \& X is Y, it is Z.} \\
{\sc All Merge} & {\it X is Y, W and Z \& X is Y in Z}  \\
{\sc "Also" cue} &  {\it X has Y, also it has Z.} \\
{\sc Distrib} &  {\it X has Y and Z.} \\
\hline
\end{tabular}
\end{footnotesize}
\caption{Scoping and Aggregation Operations in {\sc Personage} \label{aggreg-ops} }
\vspace{-.1in}
\end{table}

Aggregation describes a set of sentence planning operations that
combine multiple attributes into single sentences or phrases. We focus
here on distributive aggregation as defined in
Figure~\ref{sem-learning-fig} and illustrated in Row 6 of Table
\ref{table:agg-examples}.  In an {\sc snlg} setting, the generator
achieves this type of aggregation by operating on syntactic trees
\cite{shaw98,ScottSouza90,StentPrasadWalker04,WRR02}.  In an {\sc
  nnlg} setting, we hope the model will induce the syntactic
structure and the mathematical operation underlying it, automatically,
without explicit training supervision.

To prepare the training data, we limit the values for {\sc price} and
{\sc rating} attributes to {\sc low}, {\sc average}, and {\sc
  high}. We reserve the combination \{{\sc price=high, rating=high}\}
for test, leaving two combinations of values where distribution is
possible (\{{\sc price=low, rating=low}\} and \{{\sc price=average,
  rating=average}\}). We then use all three values in MRs where the
price and rating are not the same \{{\sc price=low, rating=high}\}.
This ensures that the model {\it does} see the value {\sc high} in
training, but never in a setting where distribution is possible. We
always distribute when possible, so every MR where the values are the
same uses distribution. All other opportunities for aggregation,
in the same sentence or in other training sentences,
use the other aggregation operations defined in {\sc personage}
as specified in Table \ref{aggreg-ops}, with equal probability.

\begin{table}[!htb]
\begin{small}
\begin{tabular}
{@{} p{0.6in}|p{0.4in}|p{0.4in}|p{0.53in}@{}}
\hline
\bf  Model & \bf Slot Error & \bf Distrib Accuracy &\bf Distrib Accuracy (on {\sc high}) \\ \midrule
{\sc  NoSup}  & .12 & 0.29       &  0.00    \\
{\sc  Binary  } & .07  & 0.99       &  0.98     \\
{\sc  Semantic  } & .25 & 0.36     &  0.09    \\ 
\bottomrule
\end{tabular}
\end{small}
\vspace{-.1in}
\centering \caption{\label{table:distrib-results} Distributive Aggregation Results}
\end{table}

The aggregation training set contains 63,690 total instances, with
19,107 instances for each of the two combinations that can distribute,
and 4,246 instances for each of the six combinations that can't
distribute. The test set contains 408 MRs, 288 specify distribution over 
{\sc high} (which we note is {\it not} a
setting seen in train, and explicitly tests the models' ability to
generalize), 30 specify distribution over {\sc average}, 30 over {\sc
  low}, and 60 are examples that do not require distribution ({\sc
  none}).  We test whether the model will learn
the equality relation independent of the value ({\sc high} vs. {\sc
  low}), and thus realize the aggregation with {\sc high}.  The
distributive aggregation experiment is based on three different
models:
\begin{itemize}  \setlength\itemsep{-0.2em}
\item {\noindent \bf No Supervision:} no additional information in the MR (only attributes and their values)
\item {\noindent \bf Binary Supervision:} we add a supervision token, {\sc distribute}, containing a binary 0 or 1 indicating whether or not the corresponding reference text contains an aggregation operation over attributes {\it price range} and {\it rating}.
\item {\noindent \bf Semantic Supervision:} we add a supervision token, {\sc distribute}, containing a string that is either {\it none} if there is no aggregation over {\it price range} and {\it rating} in the corresponding reference text, or a value of {\sc low}, {\sc average}, or {\sc high}
for aggregation. 
\end{itemize}


As above, we start
with the default TGen parameters and monitor the losses on Tensorboard
on subset of 3,000 validation instances from the 63,000 training
set. The best settings use a batch size of 20, with a minimum of 5
epochs and a maximum of 20 epochs with early-stopping.

\begin{table*}[htb!]
\begin{footnotesize}
\begin{tabular}
{@{} p{1cm}|p{6.5cm}|p{7.5cm} @{}} \toprule
\textbf{Source} & \bf MR &  \textbf{Realization}  \\ \midrule 
NYC &  name[xname], recommend[no], cuisine[xcuisine], decor[bad], qual[acceptable], location[xlocation], price[affordable], service[bad] & I imagine xname isn't great because xname is affordable, but it provides bad ambiance and rude service. It is in xlocation. It's a xcuisine restaurant with acceptable food.
\\ \hline
E2E &  name[xname], cuisine[xcuisine], location[xlocation], familyFriendly[no] & It might be okay for lunch,  but it's not a place for a family outing.
\\ \hline
E2E &  name[xname], eatType[coffee shop], cuisine[xcuisine], price[more than \$30], customerRating[low], location[xlocation], familyFriendly[yes] & Xname is a low customer rated coffee shop offering xcuisine food in the xlocation. Yes,  it is child friendly,  but the price range is more than \$30.
\\ \hline
\end{tabular}
\vspace{-.1in}
\caption{Training examples of E2E and NYC Contrast sentences}
\label{table:contrast-sentences}
\end{footnotesize}
\end{table*}

\begin{table}[!htb]
\vspace{-.1in}
\begin{scriptsize}
\begin{tabular}
{@{} p{.65in}|p{0.7in}|p{1.2in}@{}}
\hline
\bf  Training Sets & \bf NYC \#N & \bf E2E \#N \\ \midrule
{\sc \scriptsize   3K}  &    N/A    &  3,540 contrast    \\
{\sc \scriptsize  7K  } &  3,500 contrast      & 3,540 contrast    \\
{\sc \scriptsize 11K  } &  3,500 contrast    &  3,540 contrast    + 4K  random   \\ 
{\sc \scriptsize  21K  } &  3,500 contrast    &  3,540 contrast    + 14K random    \\ 
{\sc \scriptsize  21K contrast} &  3,500 contrast    &  3,540 contrast  + 14K random    \\ \hline
\end{tabular}
\end{scriptsize}
\vspace{-.1in}
\centering \caption{\label{table:train-sets-contrast} Overview of the
training sets for contrast experiments}
\end{table}




\noindent{\bf Distributive Aggregation Results.} Table
\ref{table:distrib-results} shows the accuracy of each model overall
on all 4 values, as well as the accuracy specifically on {\sc high},
the only distribution value unseen in train.  Model {\sc NoSup} has a
low overall accuracy, and is completely unable to generalize to {\sc
  high}, which is unseen in training. It is frequently able to use the
{\sc high} value, but is not able to distribute (generating output
like {\it high cost and cost}).  Model {\sc Binary} is by far the best
performing model, with an almost perfect accuracy (it is able to
distribute over {\sc low} and {\sc average} perfectly), but makes some
mistakes when trying to distribute over {\sc high}; specifically,
while it is always able to distribute, it may use an incorrect value
({\sc low} or {\sc average}). Whenever {\sc Binary} correctly
distributes over {\sc high}, it interestingly always selects attribute
{\sc rating} before {\sc cost}, realizing the output as {\it high
  rating and price}. Also, {\sc Binary} is consistent even when it
incorrectly uses the value {\sc low} instead of {\sc high}: it always
selects the attribute {\it price} before {\it rating}. To our
surprise, Model {\sc Semantic} does poorly, with 36\% overall
accuracy, and only 9\% accuracy on {\sc high}, where most of the
mistakes on {\sc high} include repeating the attribute {\it high
  rating and rating}, including examples where it does not distribute
at all, e.g. {\it high rating and high rating}. We plan to explore
alternative semantic encodings in future work.

\section{Discourse Contrast Experiment}
\label{contrast-results-sec}
 
Persuasive settings such as recommending
restaurants, hotels or travel options often have a critical discourse structure
\cite{ScottSouza90,MooreParis93,nakatsu:08}.  For example a recommendation may
consist of an explicitly evaluative utterance e.g. {\it Chanpen Thai
  is the best option}, along with content related by the justify
discourse relation, e.g. {\it It has great food and service}, as 
in Table~\ref{table:discourse-examples}. 

Our experiments focus on {\sc discourse-contrast}. We developed a
script to find contrastive sentences in the 40K E2E training set by
searching for any instance of a contrast cue word, such as {\it but},
{\it although}, and {\it even if}. This identified 3,540 instances.
While this data size is comparable to the 3-4K instances used in prior
work \cite{Wenetal15,Nayaketal17}, we anticipated that it might not be
enough data to properly test whether an {\sc nnlg} can learn to
produce discourse contrast.  We were also interested in testing
whether synthetic data would improve the ability of the {\sc nnlg} to
produce contrastive utterances while maintaining semantic
fidelity. Thus we used {\sc personage} with its native database of New
York City restaurants (NYC) to generate an additional 3,500 examples
of one form of contrast using only the discourse marker {\it but},
which are most similar to the examples in the E2E
data. Table~\ref{table:contrast-sentences} illustrates both {\sc
  personage} and E2E contrast examples. While {\sc personage} also
contains {\sc justifications}, which could possibly confuse the {\sc
  nnlg},  it  offers many more attributes that can be
contrasted and thus more unique instances of contrast. We create 4
training datasets with contrast data in order to systematically test
the effect of the combined training set.
Table~\ref{table:train-sets-contrast} provides an overview of the
training sets, with their rationales below.

\noindent{\bf 3K Training Set.} 
This dataset consists of 
all instances of contrast in the E2E training data, i.e. 3,540 E2E references.

\noindent{\bf 7K Training Set.}  We created a training set of 7k
references by supplementing the E2E contrastive references 
with an equal number of {\sc personage} references.

\noindent{\bf 11K Training Set.}  Since 7K is smaller than 
desirable for training an {\sc nnlg}, we created several additional
training sets with the aim of helping the model learn to correctly
realize domain semantics while still being able to produce contrastive
utterances.  We thus added an additional 4K crowd-sourced E2E data
that was not contrastive to our training data, for 
a total of 11,065. See Table~\ref{table:train-sets-contrast}.

\noindent{\bf 21K Training Set.} We created an additional larger
training set by adding more E2E data, again to test the effect of
increasing the size of the training set on realization of domain
semantics, without a significant decrease in our ability to produce
contrastive utterances. We added an additional 14K E2E references,
for a total of 21,065. See Table~\ref{table:train-sets-contrast}.

We perform two experiments with the 21K training set. First we
trained on the MR and reference exactly as we had done for the 7K and
11K training sets. The  second experiment 
added a contrast token during training time with values of either 1
(contrast) or 0 (no contrast) to test if that would achieve
better control of contrast.

\noindent{\bf Contrast Test Sets.} To have a potential for contrast
there must be an attribute with a positive value and another attribute
with a negative value in the same MR.  We constructed 3 different test sets, 
two for E2E and one for NYC. We created a
delexicalized version of the test set used in the E2E generation
challenge. This resulted in a test of 82 MRs of which only 25 could
support contrast ({\sc E2E Test}). In order to allow for a better test of contrast, we
constructed an additional test set of 500 E2E MRs all of which
could support contrast ({\sc E2E Contrast Test}).  For the NYC test, which provides many
opportunities for contrast, we created a dataset of 785 MRs that
were different than those seen in training ({\sc NYC Test}). 
At test time, in the 21K contrast token experiment, we utilize the
contrast token as we did in  training.

\begin{table}[!htb]
\begin{small}
\begin{tabular}
{@{} p{0.6in}||p{0.5in}|| p{0.6in}||p{0.6in} @{}}
\hline
\rowcolor[gray]{0.9}{\bf   Train} & \multicolumn{3}{c}{\bf   E2E Test (N = 82)}   \\ 
{ \cellcolor[gray]{0.9} } & { \cellcolor[gray]{0.9} {\sc Slot Errors}} & { \cellcolor[gray]{0.9} {\sc Contrast Attempts}} & {\cellcolor[gray]{0.9} {\sc Contrast Correct}} \\    \hline\hline  
{\sc   3K }  & .38  & 13    & .15  \\
{\sc   7K } & .56 & 61 &   .41  \\
{\sc   11K } & .31  & 24 & .33 \\
{\sc   21K } & .28 & 2 & .50 \\
 \hline 
{\sc   21K} &  && \\
{\sc contrast} & \bf .24 & 25  &  \bf .84\\\hline
\hline  

\end{tabular}
\end{small}
\vspace{-.1in}
\centering \caption{\label{table:E2E-results-contrast-82} Slot Error Rates and Contrast for E2E}
\end{table}

\begin{table}[!htb]
\begin{small}
\begin{tabular}
{@{} p{0.6in}||p{0.5in}|| p{0.6in}||p{0.6in} @{}}
\hline
\rowcolor[gray]{0.9}{\bf   Train} & \multicolumn{3}{c}{\bf   E2E Contrast Test (N=500)}   \\ 
{ \cellcolor[gray]{0.9} } & { \cellcolor[gray]{0.9} {\sc Slot Errors}} & { \cellcolor[gray]{0.9} {\sc Contrast Attempts}} & {\cellcolor[gray]{0.9} {\sc Contrast Correct}} \\    \hline\hline  
{\sc   3K }  & .70  & 213    & .19  \\
{\sc   7K } & .45 & 325 &   .22  \\
{\sc   11K } & .23  & 227 & .70 \\
{\sc   21K } & .17 & 13 & .62 \\
 \hline 
{\sc   21K} &  && \\
{\sc contrast} & \bf .16 & 422  &  \bf .75 \\\hline
\hline  

\end{tabular}
\end{small}
\vspace{-.1in}
\centering \caption{\label{table:E2E-results-contrast-500} Slot Error Rates and Contrast for E2E, Contrast Only}
\end{table}

\begin{table}[!htb]
\begin{small}
\begin{tabular}
{@{} p{0.6in}||p{0.5in}|| p{0.6in}||p{0.6in} @{}}
\hline

\rowcolor[gray]{0.9}{\bf   Train} & \multicolumn{3}{c}{\bf   NYC Test (N = 785)}   \\ 
{ \cellcolor[gray]{0.9} } & { \cellcolor[gray]{0.9} {\sc Slot Errors}} & { \cellcolor[gray]{0.9} {\sc Contrast Attempts}} & {\cellcolor[gray]{0.9} {\sc Contrast Correct}} \\    \hline\hline  
{\sc   3K }  & N/A  &  N/A  &  N/A \\
{\sc   7K } & .29 & 784 & .65   \\
{\sc   11K } & .26  & 696 & .71 \\
{\sc   21K } & .25 & 659 & .82 \\
 \hline 
{\sc   21K} &  && \\
{\sc contrast} & .24 & 566 & \bf .85 \\ \hline
\hline  

\end{tabular}
\end{small}
\vspace{-.1in}
\centering \caption{\label{table:NYC-results-contrast} Slot Error Rates and Contrast for NYC}
\end{table}

\noindent{\bf Contrast Results. } We present  the results for
both slot error rates and contrast  for 
the E2E test set in Table~\ref{table:E2E-results-contrast-82}, E2E Contrast in Table~\ref{table:E2E-results-contrast-500}, and NYC
test set  in Table~\ref{table:NYC-results-contrast}.

Table~\ref{table:E2E-results-contrast-82} shows the results for testing
on the original E2E test set, where we only have 25 instances 
with the possibility for contrast. 
Overall, the table shows 
large performance improvements with the {\sc contrast}
token supervision for 21K for both slot errors
and correct contrast. On the  E2E test set, the 
the 3K E2E training set gives
a slot error rate of .38 and only 15\% correct contrast. 
The 7K training set, supplemented with additional generated
contrast examples gets a correct contrast of .41 but a much higher
slot error rate. Interestinglyx, the 11K dataset is much
better than the 3K for contrast correct, suggesting a positive
effect for the automatically generated contrast examples along
with more E2E training data. The 21K set without the contrast
token does not attempt contrast since the frequency of contrast
data is low, but with the {\sc contrast} token, it attempts
contrast every time it is possible (25/25 instances).

In Table~\ref{table:E2E-results-contrast-500} with only contrast data, we see similar trends, with the 
lowest slot error rate (.16) and highest correct contrast (.75) ratios for the experiment 
with token supervision on 21K. Again, we see 
much better performance from the 11K set than the 3K and 7K in terms of slot error 
and correct contrast, indicating that more training data (even if that data does not contain contrast) helps the model. As before, we see very low contrast attempts with 21K without supervision, with a huge increase in the number of contrast attempts when using token supervision (422/500).

Table~\ref{table:NYC-results-contrast} also shows large performance
improvements from the use of the {\sc contrast} token supervision for
the NYC test set, again with improvements for the 21K {\sc contrast} in
both slot error rate and in correct contrast. Interestingly, while we
get the highest correct contrast ratio of .85 with {\sc 21K Contrast},
we actually see {\it fewer} contrast attempts, showing that the most
explicitly supervised model is becoming more selective when deciding
when to do contrast. When training on the 7K dataset, the neural model
{\bf always} produces a contrastive utterance for the NYC MRs (all the
NYC data is contrastive).  Although it never sees any NYC
non-contrastive MRs, the additional E2E training data allows it to
improve its ability to decide when to contrast (Row {\sc 21K
  contrast}) as well as improving the slot error rate in the final
experiment.




\section{Related Work}

Much of the previous work focused on sentence planning was done in the
framework of statistical {\sc nlg}, where each module was assumed to
require training data that matched its representational
requirements. Methods focused on training individual modules for
content selection and linearization
\cite{Marcu97,Lapata2003,BarzilayLapata05}, and trainable sentence
planning for discourse structure and aggregation operations
\cite{stentmolina09,Walkeretal07,PaivaEvans2004,SauperBarzilay09,ChengPoesio}.
Previous work also explored statistical and hybrid methods for surface
realization
\cite{LangkildeKnight98,BangaloreRambow2000,ohrudnicky02}.
and text-to-speech realizations
\cite{Hitzeman98,BulykoOstendorf01,Hirschberg93}.

Other work on {\sc nnlg} also uses token supervision and modifications
of the architecture in order to control stylistic aspects of the
output in the context of text-to-text or paraphrase generation. Some
types of stylistic variation correspond to sentence planning
operations, e.g. to express a particular personality type
\cite{Orabyetal18,MairesseWalker11,oraby2018neural}, or to control
sentiment and sentence theme \cite{Ficler17}. Herzig et
al. \shortcite{Herzig17} automatically label the personality of
customer care agents and then control the personality during
generation.  Rao and Tetreault \shortcite{RaoTetreault18} train a
model to paraphrase from formal to informal style and Niu and Bansal
\shortcite{NiuBansal18} use a high precision classifier and a blended
language model to control utterance politness.

Previous work on contrast has explored how the user model determines
which values should be contrasted, since people may have differing
opinions about whether an attribute value is positive or negative
(e.g. {\it family friendly})
\cite{CareniniMoore93,Walkeretal02,White-Clark-Moore:2010}. To our
knowledge, no-one has yet trained an {\sc nnlg} to use a model of user
preferences for content selection. Here, values are treated as
inherently good or bad, e.g.  service is ranked from great to
terrible.

\section{Discussion and Conclusion}
\label{conc-sec}

This paper presents detailed, systematic experiments to test the
ability of {\sc nnlg} models to produce complex sentence planning
operations for response generation. We create new training and test
sets designed specifically for testing sentence planning operations
for sentence scoping, aggregation and discourse contrast, and train
novel models with increasing levels of supervision to examine how much
information is required to control neural sentence planning. The
results show that the models benefit from extra latent variable
supervision, which improves the {\bf semantic accuracy} of the {\sc
  nnlg}, provides the capability to {\bf control} variation in the
output, and enables {\bf generalizing} to unseen value combinations.

In future work we plan to test these methods in different domains,
e.g.  the WebNLG challenge or WikiBio dataset
\cite{Wiseman2018LearningNT,colin2016webnlg}. We also plan to
experiment with more complex sentence planning operations and test
whether an {\sc nnlg} system can be endowed with fine-tuned control,
e.g. controlling multiple aggregation operations.  Another possibility
is that hierarchical input representations representing the sentence
plan might improve performance or allow finer-grained control
\cite{FLIGHTS-FLAIRS:2004,su2018investigating,BangaloreRambow2000}.
It may be desirable to control which attributes are aggregated
together, distributed or contrasted, and to allow more than two values
to be contrasted.

Here, our main goal was to test the ability of different neural
architectures to learn particular sentence planning operations that
have been used in previous work in {\sc snlg}.  Because we don't make
claims about fluency or naturalness, we did not evaluate these with
human judgements. Instead, we focused our evaluation on automatic
assessment of semantic fidelity, and the extent to which the neural
architecture could reproduce the desired sentence planning operations.
In future work, we hope to quantify the extent to which human subjects
prefer the outputs where the sentence planning operations have been
applied.

\section{Acknowledgments} 
This work was supported by NSF Cyberlearning EAGER grant IIS 1748056 
and NSF Robust Intelligence IIS 1302668-002 as well as an Amazon Alexa
Prize Gift 2017 and Grant 2018 awarded to the Natural Language
and Dialogue Systems Lab at UCSC.

\bibliography{../../../nl,../../../phd}
\bibliographystyle{acl_natbib}

\end{document}